\pdfoutput=1
\documentclass[11pt]{article}
\usepackage[final]{acl}

\usepackage{times}
\usepackage{latexsym}
\usepackage{listings}
\usepackage{geometry}
\usepackage{color}
\usepackage{colortbl}

\usepackage[T1]{fontenc}
\usepackage[utf8]{inputenc}
\usepackage{microtype}
\usepackage{inconsolata}

\usepackage{url}
\usepackage{amsmath}
\usepackage{amssymb}
\usepackage{graphicx}
\usepackage[normalem]{ulem}
\usepackage{multirow}
\usepackage{booktabs}
\usepackage[cjk]{kotex}
\usepackage{makecell}

\usepackage{CJKpunct}

\newcommand{\std}[1]{\text{\scriptsize #1}}

\usepackage{CJKutf8}

\usepackage{array}
\usepackage{float}

\newcommand\blfootnote[1]{%
  \begingroup
  \renewcommand\thefootnote{}\footnote{#1}%
  \addtocounter{footnote}{-1}%
  \endgroup
}

\title{Cross-Lingual Optimization for Language Transfer in Large Language Models}

\author{
  Jungseob Lee$^{*}$, 
  Seongtae Hong$^{*}$, 
  Hyeonseok Moon, 
  Heuiseok Lim$^{\dagger}$ \\
  Korea University \\
  \texttt{\{omanma1928, ghdchlwls123, glee889, limhseok\}@korea.ac.kr}
}

\begin{document}
\maketitle
\begin{abstract}
\blfootnote{This paper has been accepted at the ACL 2025 main conference.}
\blfootnote{$^*$ Equal contribution; $^\dagger$ Corresponding author}
Adapting large language models to other languages typically employs supervised fine-tuning (SFT) as a standard approach. However, it often suffers from an overemphasis on English performance, a phenomenon that is especially pronounced in data-constrained environments. To overcome these challenges, we propose \textbf{Cross-Lingual Optimization (CLO)} that efficiently transfers an English-centric LLM to a target language while preserving its English capabilities. CLO utilizes publicly available English SFT data and a translation model to enable cross-lingual transfer. We conduct experiments using five models on six languages, each possessing varying levels of resource. Our results show that CLO consistently outperforms SFT in both acquiring target language proficiency and maintaining English performance. Remarkably, in low-resource languages, CLO with only 3,200 samples surpasses SFT with 6,400 samples, demonstrating that CLO can achieve better performance with less data. Furthermore, we find that SFT is particularly sensitive to data quantity in medium and low-resource languages, whereas CLO remains robust. Our comprehensive analysis emphasizes the limitations of SFT and incorporates additional training strategies in CLO to enhance efficiency.
\end{abstract}

\section{Introduction}
\begin{figure}[t]
 \centering
\includegraphics[width=1.0\linewidth]{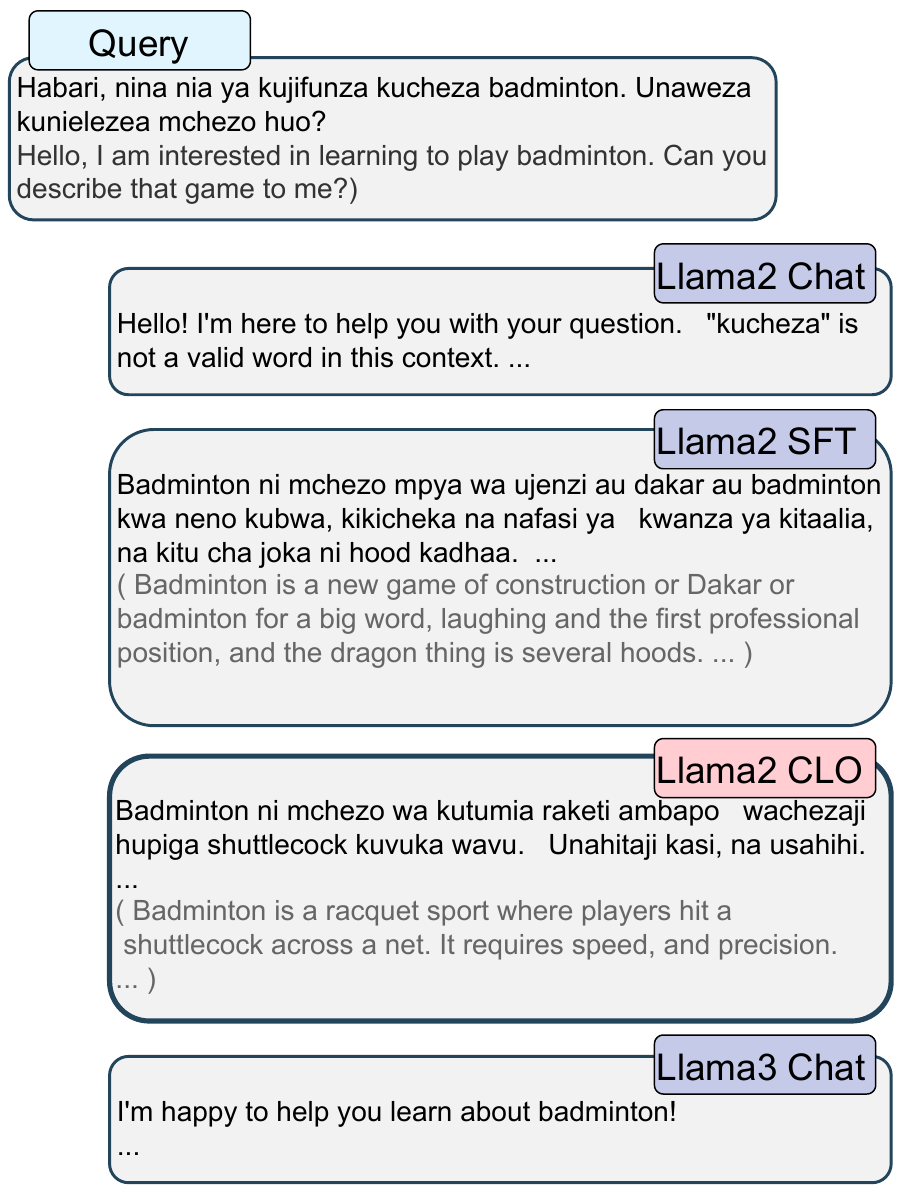}
\caption{Example responses to a Swahili query generated by English-centric instruction models, the SFT model, and the proposed CLO model.}
 \label{fig:example_intro}
\end{figure}

While the rapid advancement of Large Language Models (LLMs) has led to significant innovations in natural language processing \cite{achiam2023gpt, waisberg2023gpt}, these models are primarily pre-trained on English data and consequently exhibit relatively lower performance for other languages \cite{touvron2023llama2, dubey2024llama_, team2024gemma}. This discrepancy largely stems from the data distribution imbalance and scarcity of the pre-training and alignment data~\cite{muraoka2023cross}.

We identify the limited multilingual capabilities of English-centric language models in the following two cases. First, these models may fail to comprehend certain languages. Second, they might understand a language but still default to communicating in English~\cite{marchisio2024understanding}. Figure~\ref{fig:example_intro} illustrates these scenarios. When posed with the Swahili question, the Llama2 Chat model~\cite{touvron2023llama2} fails to comprehend Swahili adequately, while the Llama3 Chat model \cite{dubey2024llama_} understands the query but is unable to generate responses in Swahili, opting for English instead. We notice that even the fine-tuned model, despite being trained with 6,400 Swahili data, still struggles to produce appropriate outputs in Swahili~\cite{zhao2024llama, chirkova2024zero}.

Based on these observations, we identified that standard fine-tuning (SFT) methods struggle to achieve direct alignment with English in data scarcity scenario. Then we hypothesize that simultaneously enhancing target language ability and aligning it with English will facilitate efficient transfer. In this context, we suggest a cross-lingual response prioritization method to strengthen the target language ability while aligning English and the target language. By promoting the preference for responding in the target language when provided with inputs in that same language, we argue that it is possible to utilize the model’s embedded language knowledge while preserving its existing English capabilities.

To address this challenge, we propose the Cross-Lingual Optimization (CLO) strategy. CLO aims to effectively transfer an LLM to a target language using translated data. Specifically, it modifies the Direct Preference Optimization (DPO)~\cite{rafailov2024direct} approach to increase the preference for responding in the same language as the input. Simultaneously, it reduces the preference for responding in different languages for a given input, thereby facilitating knowledge acquisition in the target language.

To validate the effectiveness of CLO, we experiment with a resources-limited environment. In this process, we used 6,400 publicly available English seed data points and an accessible translation model to target languages, along with five LLMs. Through comparative experiments with traditional transfer methods, we demonstrate superiority of the CLO method in terms of instruction-following ability and NLP benchmarks. Our results show that SFT exhibits better target language adaptation in high-resource settings, such as Chinese, while in low-resource languages like Swahili it tends to over-prioritize English. In contrast, our CLO approach demonstrated consistent performance improvements across all languages and models in both the target language and English. Furthermore, motivated by recent findings, we explored targeted fine-tuning strategies that reliably matched the performance of full model training. Accordingly, we adopt this training method as our primary strategy for facilitating effective language adaptation.

\begin{figure*}[t]
 \centering
\includegraphics[width=1.0\linewidth]{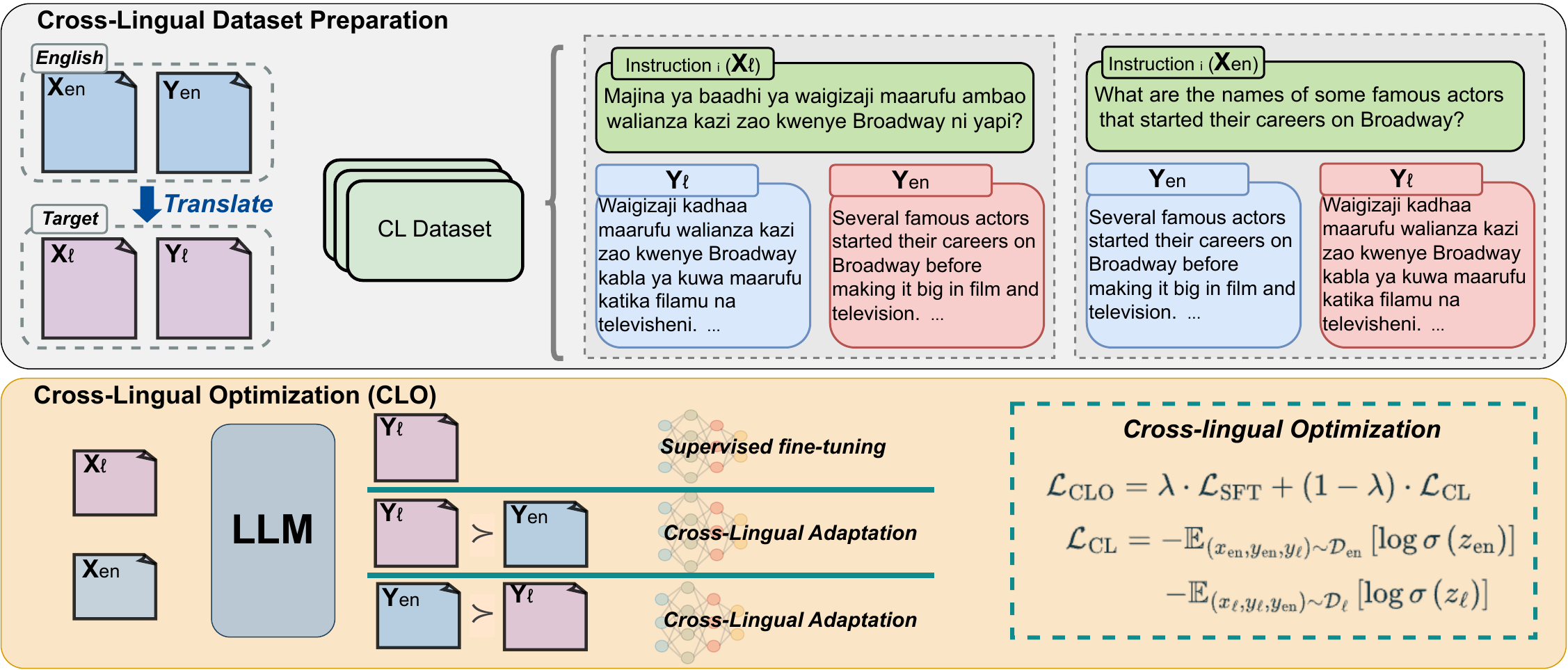}
\caption{Overview of cross-lingual dataset preparation and optimization method. The process begins with translating English (\(x_{\text{en}}\), \(y_{\text{en}}\)) pairs into a target language to create a cross-lingual dataset. This process results in the creation of (\(x_{\ell}\), \(y_{\ell}\)) pairs in the target language. The optimization is performed using a combined loss \(\mathcal{L}_{\text{CLO}}\).}
 \label{fig:CLO}
\end{figure*}

\section{Related Works}
Research on enhancing performance by transferring English-centric pre-trained language models to other languages is actively ongoing~\cite{tran2020english, minixhofer2021wechsel, de2021adapting, li2024dynamic, chen2024instructioncp}. These studies primarily explore how effectively models trained in English can transfer with minimal data~\cite{dobler2023focus}, as well as their ability to follow instructions in other languages~\cite{zhao2024llama}.

Most language transfer studies either train English-centric LLMs with instruction tuning data in the target language~\cite{lee2023kullm, shaham2024multilingual} or perform instruction tuning after continual pre-training on a large corpus in the target language~\cite{cui2023efficient, zhao2024llama}. However, many of these language transfer studies often require large-scale training datasets \cite{chen2024instructioncp, li2024dynamic} or involve complex model architecture analyses \cite{zhao2024adamergex, lee2024length}.

Inspired by \citet{shaham-etal-2024-multilingual}, which demonstrated high cross-lingual performance with a limited number of multilingual examples, we explore strategies for achieving effective transfer with small-sized datasets. In this paper, we propose an efficient language transfer method that addresses real-world scenarios, such as those in low-resource languages where instruction tuning data is unavailable, by leveraging publicly available translation models as a component and adapting an English-centric LLM to the target language using only English instruction tuning data.

\section{Our Frameworks}
Our cross-lingual transfer method assumes the availability of a base language model, a small amount of unidirectional English SFT data, and a translation model that supports the target language. The first key hypothesis is that given an input query in a non-English target language, suppressing English responses while strengthening responses in the target language enables the model to leverage its existing English knowledge to generate outputs in the target language. The second key hypothesis is that to transfer the ability to generate responses in English to the target language, it is sufficient to include a relatively small number of target language responses in a consistent response format. 

Based on these hypotheses, we modify the Direct Preference Optimization (DPO) loss \cite{rafailov2024direct} to suit our purpose. Since the base language model does not initially answer the queries, we combine the Negative Log-Likelihood (NLL) with the modified DPO loss to produce appropriate responses. Our process consists of two steps:
\begin{enumerate}
    \item \textbf{Cross-Lingual Dataset Preparation}: Obtain translated data from the original English preference dataset.
    \item \textbf{Cross-Lingual Optimization}: Train the attention layers using the translated data to generate responses in the query language.
\end{enumerate}
Figure~\ref{fig:CLO} illustrates our approach, and we detail the proposed CLO method below.

\subsection{Brief Overview of Standard DPO}
DPO employs an analytical mapping from a reward function to derive an optimal policy without the need for an explicit reward model. The standard DPO loss is defined as:
\begin{equation}
\begin{aligned}
\mathcal{L}_{\text{DPO}}(\pi_\theta; \pi_{\text{ref}}) = 
- \mathbb{E}_{\left(\mathbf{x}, \mathbf{y}_w, \mathbf{y}_l\right) \sim \mathcal{D}} \left[ \log \sigma \left( z \right) \right]
\end{aligned}
\end{equation}
where \( z \) is defined as:
\begin{equation}
\begin{aligned}
z = 
\beta \log \frac{\pi_\theta(y_w \mid x)}{\pi_{\text{ref}}(y_w \mid x)}
- \beta \log \frac{\pi_\theta(y_l \mid x)}{\pi_{\text{ref}}(y_l \mid x)}
\end{aligned}
\end{equation}

where \(\pi_\theta\) is the policy, \(\pi_{\text{ref}}\) the reference policy, \(\sigma\) represents the sigmoid function, and \(\beta\) regulates the KL constraint.

\subsection{CLO Methodology}
\paragraph{Cross-Lingual Dataset Preparation}
We generate target language prompts \(\mathbf{x}_{\ell}\) and responses \(\mathbf{y}_{\ell}\) by translating the English prompts \(\mathbf{x}_{\text{en}}\) and responses \(\mathbf{y}_{\text{en}}\) in the existing SFT dataset using a translation model.

For \(\mathbf{x}_{\text{en}}\), we associate the \emph{chosen response} with \(\mathbf{y}_{\text{en}}\) and the \emph{rejected response} with \(\mathbf{y}_{\ell}\) to form the English training data. This mapping prevents the model from generating target language outputs when provided an English prompt, thereby preserving its English capability. Conversely, for \(\mathbf{x}_{\ell}\), we map the \emph{chosen response} to \(\mathbf{y}_{\ell}\) and the \emph{rejected response} to \(\mathbf{y}_{\text{en}}\) to form the target language data. This encourages the model, when given a target language instruction, to generate responses in the target language by relying on its underlying English knowledge.

\paragraph{Cross-Lingual Optimization}
Our proposed CLO introduces a new loss function that differs from the standard DPO loss by utilizing cross-lingual data pairs within the \emph{same} batch to explicitly teach the model the correspondence between input and output languages. Importantly, to mitigate inherent English bias, we only consider the NLL loss computed on the target language outputs. Further, Based on the findings of \citet{anonymous2024neuronlevel}, which highlight the critical role of attention layers in language capabilities, we accordingly fine-tune only the attention layers. The overall loss function of CLO is defined as:
\begin{equation}
\begin{aligned}
\mathcal{L}_{\text{CLO}} = \lambda \cdot \mathcal{L}_{\text{SFT}} + (1-\lambda) \cdot \mathcal{L}_{\text{CL}}
\end{aligned}
\end{equation}
where \(\mathcal{L}_{\text{SFT}}\) is the standard NLL loss over the target language data \(\mathcal{D}_{\ell}\):
\begin{equation}
\begin{aligned}
\mathcal{L}_{\text{SFT}} = \mathbb{E}_{\left( \mathbf{x}_{\ell}, \mathbf{y}_{\ell} \right) \sim \mathcal{D}_{\ell}} \left[
 - \log \pi_{\theta_{\text{att}}}(y_\ell \mid x_\ell) \right]
\end{aligned}
\end{equation}
and \(\mathcal{L}_{\text{CL}}\) is the cross-lingual loss that explicitly enforces language correspondence, defined as:
\begin{equation}
\begin{aligned}
\mathcal{L}_{\text{CL}} = 
& -\mathbb{E}_{\left( \mathbf{x}_{\text{en}}, \mathbf{y}_{\text{en}}, \mathbf{y}_{\ell} \right) \sim \mathcal{D}_{\text{en}}} \left[ \log \sigma \left( z_{\text{en}} \right) \right] \\
& -\mathbb{E}_{\left( \mathbf{x}_{\ell}, \mathbf{y}_{\ell}, \mathbf{y}_{\text{en}} \right) \sim \mathcal{D}_{\ell}} \left[ \log \sigma \left( z_{\ell} \right) \right]
\end{aligned}
\end{equation}
where \( z_{\text{en}} \) and \( z_{\ell} \) are defined respectively as:
\begin{equation}
\begin{aligned}
z_{\text{en}} = \beta \Biggl( \log\!\frac{\pi_{\theta_{\text{att}}}(y_{\text{en}} \mid x_{\text{en}})}
{\pi_{\text{ref}}(y_{\text{en}} \mid x_{\text{en}})} 
- \log\!\frac{\pi_{\theta_{\text{att}}}(y_{\ell} \mid x_{\text{en}})}
{\pi_{\text{ref}}(y_{\ell} \mid x_{\text{en}})} \Biggr)
\end{aligned}
\end{equation}
\begin{equation}
\begin{aligned}
z_{\ell} = \beta \left( \log \frac{\pi_{\theta_{\text{att}}}(y_{\ell} \mid x_{\ell})}{\pi_{\text{ref}}(y_{\ell} \mid x_{\ell})} - \log \frac{\pi_{\theta_{\text{att}}}(y_{\text{en}} \mid x_{\ell})}{\pi_{\text{ref}}(y_{\text{en}} \mid x_{\ell})} \right)
\end{aligned}
\end{equation}
By incorporating both preferred and rejected responses in different languages within the same batch, the model is guided to generate outputs in the appropriate language based on the input, effectively transferring its English knowledge to the target language while maintaining proficiency in English. The differences between the standard DPO loss and the Cross-lingual loss are detailed in Appendix~\ref{appendix:cross-lingual-loss}.

\begin{table*}[ht]
\centering
\resizebox{1.0\linewidth}{!}{
\begin{tabular}{%
    ll
    ccc
    ccc
    ccc
    ccc
    ccc
    ccc
}
\toprule
& & \multicolumn{6}{c}{High-Resource} 
  & \multicolumn{6}{c}{Medium-Resource}
  & \multicolumn{6}{c}{Low-Resource} \\
\cmidrule(lr){3-8}\cmidrule(lr){9-14}\cmidrule(lr){15-20}

\multirow{2}{*}{Model} & \multirow{2}{*}{Eval Lang} 
& \multicolumn{3}{c}{Chinese} 
& \multicolumn{3}{c}{German} 
& \multicolumn{3}{c}{Korean}
& \multicolumn{3}{c}{Indonesian}
& \multicolumn{3}{c}{Swahili} 
& \multicolumn{3}{c}{Yoruba}
\\
\cmidrule(lr){3-5}
\cmidrule(lr){6-8}
\cmidrule(lr){9-11}
\cmidrule(lr){12-14}
\cmidrule(lr){15-17}
\cmidrule(lr){18-20}
& & SFT+DPO & CLO & $\Delta$
  & SFT+DPO & CLO & $\Delta$
  & SFT+DPO & CLO & $\Delta$
  & SFT+DPO & CLO & $\Delta$
  & SFT+DPO & CLO & $\Delta$
  & SFT+DPO & CLO & $\Delta$
\\
\midrule
\multirow{2}{*}{\textbf{Llama-3-8B}} 
& Target
& $59.1$\std{$\pm1.70$} & $70.4$\std{$\pm1.62$} & \cellcolor[HTML]{FFABAB}\textbf{+11.3}
& $51.7$\std{$\pm1.76$} & $54.6$\std{$\pm1.76$} & \cellcolor[HTML]{FFECEC}\textbf{+2.9}
& $62.5$\std{$\pm1.70$} & $77.8$\std{$\pm1.47$} & \cellcolor[HTML]{FF8A80}\textbf{+15.3}
& $54.9$\std{$\pm1.75$} & $56.4$\std{$\pm1.75$} & \cellcolor[HTML]{FFECEC}\textbf{+1.5}
& $65.4$\std{$\pm1.67$} & $83.0$\std{$\pm1.61$} & \cellcolor[HTML]{FF8A80}\textbf{+17.6}
& $52.4$\std{$\pm1.76$} & $64.0$\std{$\pm1.70$} & \cellcolor[HTML]{FFABAB}\textbf{+11.6}
\\
& English
& $52.0$\std{$\pm1.76$} & $64.0$\std{$\pm1.73$} & \cellcolor[HTML]{FFABAB}\textbf{+12.0}
& $50.1$\std{$\pm1.76$} & $55.5$\std{$\pm1.76$} & \cellcolor[HTML]{FFD7D5}\textbf{+5.4}
& $52.1$\std{$\pm1.76$} & $64.4$\std{$\pm1.69$} & \cellcolor[HTML]{FFABAB}\textbf{+12.3}
& $51.6$\std{$\pm1.76$} & $57.7$\std{$\pm1.75$} & \cellcolor[HTML]{FFD7D5}\textbf{+6.1}
& $50.3$\std{$\pm1.76$} & $61.4$\std{$\pm1.73$} & \cellcolor[HTML]{FFABAB}\textbf{+11.1}
& $51.2$\std{$\pm1.76$} & $58.2$\std{$\pm1.75$} & \cellcolor[HTML]{FFD7D5}\textbf{+7.0}
\\
\midrule
\multirow{2}{*}{\textbf{Llama-2-7B}} 
& Target
& $59.1$\std{$\pm1.73$} & $61.1$\std{$\pm1.72$} & \cellcolor[HTML]{FFECEC}\textbf{+2.0}
& $50.3$\std{$\pm1.76$} & $59.5$\std{$\pm1.73$} & \cellcolor[HTML]{FFD7D5}\textbf{+9.2}
& $52.5$\std{$\pm1.76$} & $53.8$\std{$\pm1.76$} & \cellcolor[HTML]{FFECEC}\textbf{+1.3}
& $50.8$\std{$\pm1.76$} & $55.8$\std{$\pm1.75$} & \cellcolor[HTML]{FFD7D5}\textbf{+5.0}
& $64.1$\std{$\pm1.64$} & $65.0$\std{$\pm1.72$} & \cellcolor[HTML]{FFECEC}\textbf{+0.9}
& $43.5$\std{$\pm1.74$} & $67.1$\std{$\pm1.66$} & \cellcolor[HTML]{FF8A80}\textbf{+23.6}
\\
& English
& $51.1$\std{$\pm1.76$} & $55.7$\std{$\pm1.76$} & \cellcolor[HTML]{FFD7D5}\textbf{+4.6}
& $48.4$\std{$\pm1.76$} & $61.2$\std{$\pm1.72$} & \cellcolor[HTML]{FFABAB}\textbf{+12.8}
& $50.6$\std{$\pm1.76$} & $51.0$\std{$\pm1.76$} & \cellcolor[HTML]{FFECEC}\textbf{+0.4}
& $49.3$\std{$\pm1.76$} & $62.4$\std{$\pm1.71$} & \cellcolor[HTML]{FFABAB}\textbf{+13.1}
& $51.5$\std{$\pm1.76$} & $55.5$\std{$\pm1.75$} & \cellcolor[HTML]{FFD7D5}\textbf{+4.0}
& $50.3$\std{$\pm1.76$} & $61.2$\std{$\pm1.72$} & \cellcolor[HTML]{FFABAB}\textbf{+10.9}
\\
\midrule
\multirow{2}{*}{\textbf{Llama-2-13B}}
& Target
& $59.3$\std{$\pm1.74$} & $65.2$\std{$\pm1.71$} & \cellcolor[HTML]{FFD7D5}\textbf{+5.9}
& $50.5$\std{$\pm1.76$} & $53.7$\std{$\pm1.76$} & \cellcolor[HTML]{FFECEC}\textbf{+3.2}
& $51.6$\std{$\pm1.75$} & $53.9$\std{$\pm1.75$} & \cellcolor[HTML]{FFECEC}\textbf{+2.3}
& $52.4$\std{$\pm1.76$} & $61.7$\std{$\pm1.71$} & \cellcolor[HTML]{FFD7D5}\textbf{+9.3}
& $53.9$\std{$\pm1.56$} & $70.9$\std{$\pm1.60$} & \cellcolor[HTML]{FF8A80}\textbf{+17.0}
& $43.5$\std{$\pm1.75$} & $67.3$\std{$\pm1.65$} & \cellcolor[HTML]{FF8A80}\textbf{+23.8}
\\
& English
& $50.6$\std{$\pm1.76$} & $50.4$\std{$\pm1.76$} & \cellcolor[HTML]{C9E4FF}\textbf{-0.2}
& $53.0$\std{$\pm1.76$} & $58.5$\std{$\pm1.74$} & \cellcolor[HTML]{FFD7D5}\textbf{+5.5}
& $51.4$\std{$\pm1.76$} & $52.3$\std{$\pm1.76$} & \cellcolor[HTML]{FFECEC}\textbf{+0.9}
& $52.9$\std{$\pm1.76$} & $59.8$\std{$\pm1.73$} & \cellcolor[HTML]{FFD7D5}\textbf{+6.9}
& $50.8$\std{$\pm1.76$} & $55.0$\std{$\pm1.75$} & \cellcolor[HTML]{FFD7D5}\textbf{+4.2}
& $51.1$\std{$\pm1.76$} & $54.2$\std{$\pm1.76$} & \cellcolor[HTML]{FFECEC}\textbf{+3.1}
\\
\midrule
\multirow{2}{*}{\textbf{Mistral-7B-v0.1}} 
& Target
& $56.8$\std{$\pm1.75$} & $57.4$\std{$\pm1.74$} & \cellcolor[HTML]{FFECEC}\textbf{+0.6}
& $49.9$\std{$\pm1.76$} & $50.8$\std{$\pm1.76$} & \cellcolor[HTML]{FFECEC}\textbf{+0.9}
& $48.5$\std{$\pm1.76$} & $50.5$\std{$\pm1.76$} & \cellcolor[HTML]{FFECEC}\textbf{+2.0}
& $50.0$\std{$\pm1.76$} & $51.1$\std{$\pm1.76$} & \cellcolor[HTML]{FFECEC}\textbf{+1.1}
& $35.4$\std{$\pm1.71$} & $51.3$\std{$\pm1.76$} & \cellcolor[HTML]{FF8A80}\textbf{+15.9}
& $50.9$\std{$\pm1.76$} & $51.1$\std{$\pm1.52$} & \cellcolor[HTML]{FFECEC}\textbf{+0.2}
\\
& English
& $51.4$\std{$\pm1.76$} & $57.1$\std{$\pm1.75$} & \cellcolor[HTML]{FFD7D5}\textbf{+5.7}
& $48.6$\std{$\pm1.76$} & $50.6$\std{$\pm1.76$} & \cellcolor[HTML]{FFECEC}\textbf{+2.0}
& $52.9$\std{$\pm1.76$} & $65.2$\std{$\pm1.71$} & \cellcolor[HTML]{FFABAB}\textbf{+12.3}
& $52.3$\std{$\pm1.76$} & $54.8$\std{$\pm1.76$} & \cellcolor[HTML]{FFECEC}\textbf{+2.5}
& $51.2$\std{$\pm1.76$} & $64.2$\std{$\pm1.69$} & \cellcolor[HTML]{FFABAB}\textbf{+13.0}
& $49.4$\std{$\pm1.76$} & $52.8$\std{$\pm1.76$} & \cellcolor[HTML]{FFECEC}\textbf{+3.4}
\\
\midrule
\multirow{2}{*}{\textbf{Qwen-2.5-3B}}
& Target
& $54.2$\std{$\pm1.75$} & $59.9$\std{$\pm1.73$} & \cellcolor[HTML]{FFD7D5}\textbf{+5.7}
& $53.0$\std{$\pm1.76$} & $54.0$\std{$\pm1.76$} & \cellcolor[HTML]{FFECEC}\textbf{+1.0}
& $52.3$\std{$\pm1.76$} & $62.4$\std{$\pm1.71$} & \cellcolor[HTML]{FFABAB}\textbf{+10.1}
& $53.8$\std{$\pm1.76$} & $54.9$\std{$\pm1.75$} & \cellcolor[HTML]{FFECEC}\textbf{+1.1}
& $51.7$\std{$\pm1.76$} & $74.7$\std{$\pm1.53$} & \cellcolor[HTML]{FF8A80}\textbf{+23.0}
& $45.9$\std{$\pm1.76$} & $68.9$\std{$\pm1.63$} & \cellcolor[HTML]{FF8A80}\textbf{+23.0}
\\
& English
& $50.7$\std{$\pm1.76$} & $56.7$\std{$\pm1.75$} & \cellcolor[HTML]{FFD7D5}\textbf{+6.0}
& $51.9$\std{$\pm1.76$} & $54.8$\std{$\pm1.76$} & \cellcolor[HTML]{FFECEC}\textbf{+2.9}
& $50.8$\std{$\pm1.76$} & $61.0$\std{$\pm1.72$} & \cellcolor[HTML]{FFABAB}\textbf{+10.2}
& $50.6$\std{$\pm1.76$} & $58.0$\std{$\pm1.75$} & \cellcolor[HTML]{FFD7D5}\textbf{+7.4}
& $51.1$\std{$\pm1.76$} & $58.1$\std{$\pm1.74$} & \cellcolor[HTML]{FFD7D5}\textbf{+7.0}
& $54.4$\std{$\pm1.76$} & $57.0$\std{$\pm1.75$} & \cellcolor[HTML]{FFECEC}\textbf{+2.6}
\\
\bottomrule
\end{tabular}
}
\caption{Win-rate (\%) results on AlpacaEval for models fine-tuned with SFT+DPO and CLO, evaluated against their SFT baselines. Each cell reports the win rate and its standard deviation. The $\Delta$ denotes the absolute improvement of CLO over SFT+DPO.}
\label{tab:alpaca}
\end{table*}

\section{Experimental Setup}
\subsection{Training Setup}

We primarily compare our CLO method with the standard SFT baseline. In some experiments, we also include results for models trained with SFT followed by DPO, following the experimental protocol described by \citet{hong2024orpo}. Specifically, \textit{SFT+DPO} refers to a two-stage training procedure in which SFT is first performed using the preferred responses for each data point, and then DPO is applied to the same dataset. This approach evaluates the effect of simple data augmentation using cross-lingual datasets. The experimental results indicate that using cross-lingual datasets without the CLO objective function does not achieve strong language transfer performance.

We use 6,400 data samples from OpenAssistant~\cite{kopf2024openassistant}, selected in order of highest ranking\footnote{The dataset provides a \texttt{rank} column, which indicates the relative quality or preference order of assistant responses to a given prompt, as judged by human annotators.}, as our English single-turn dataset. Each example is a selected (instruction, output) pair from the first turn.

To validate the effectiveness of CLO, we conduct experiments on six languages with varying resource availability, using five available pre-trained LLMs: Llama-2-7B, Llama-2-13B~\cite{touvron2023llama}, Llama-3-8B~\cite{dubey2024llama_}, Mistral-7B-v0.1~\cite{jiang2023mistral}, and Qwen-2.5-3B~\cite{yang2024qwen2}. Specifically, we choose two high-resource languages (Chinese, German), two medium-resource languages (Korean, Indonesian), and two low-resource languages (Swahili, Yoruba) based on the amount of pre-training corpus data in the Llama-2~\cite{touvron2023llama}. For each of the specified target languages, we create cross-lingual data by translating all the extracted English (instruction, output) pairs \(\{\left(\mathbf{x}_{\text{en}}, \mathbf{y}_{\text{en}}\right)\}\) into \(\{\left(\mathbf{x}_{\text{en}}, \mathbf{y}_{\text{en}}, \mathbf{x}_{\ell}, \mathbf{y}_{\ell}\right)\}\).

We adopt the M2M100 1.2B translation model~\cite{fan2021beyond} to generate a total of 12,800 samples across the languages (6,400 in English and 6,400 in the target language). We trained all baselines and models using identical hyperparameters and conducted evaluations under the same settings. The details of the hyperparameters used during training and generation are provided in Appendix~\ref{appendix:hyperparameters}.

\subsection{Evaluation Benchamrk Setup}

\paragraph{AlpacaEval}
To investigate the instruction-following ability of the models, we use AlpacaEval~\cite{alpaca_eval}, translating it into each respective language, and measure the win rate against the baseline SFT model. We include in the evaluation whether the model responded appropriately in the given language, in addition to the original prompts from AlpacaEval, and use the GPT-4o\footnote{\url{https://openai.com/index/hello-gpt-4o/}, \textit{gpt-4o-2024-08-06}} for comparisons. The evaluation prompts and detailed information are provided in Appendix~\ref{appen:alpacaeval}.

\paragraph{MRC Benchmark}
To assess the machine reading comprehension (MRC) capabilities of the models, we employ the BELEBELE dataset~\cite{bandarkar2023belebele}. In this benchmark, each instance consists of a question and a passage accompanied by four answer choices. Each instance features human-curated questions and answers that are designed to be challenging by including plausible distractors. The model’s accuracy is determined by computing the log likelihood of each answer option and selecting the one with the highest overall likelihood~\cite{eval-harness}.

To support our multilingual experiments, we use language-specific prompt templates that maintain a consistent structure across languages while using native terms for key input fields. In these templates, the input fields are mapped to their corresponding local labels. Table~\ref{tab:prompt_templates_belebele} describes these mappings for each target language, including the appropriate answer indicator. In our experiments, each formatted prompt is passed as a single text string into the model to evaluate its zero-shot performance across six languages.

\paragraph{MMMLU}
To measure the models' reasoning performance, we utilize OpenAI's Multilingual Massive Multi-task Language Understanding\footnote{\url{https://github.com/openai/simple-evals}} (MMMLU)~\cite{hendryckstest2021}. This test set, developed with the input of professional human translators, extracts answers not based on the probabilities of the correct tokens but from the model's generated outputs. Since the model infers answers based on the step-by-step reasoning path, it allows for more accurate measurements than traditional token probabilities. Additionally, because the test includes instructions that require the model to respond in specific answer formats, the model's instruction-following ability is essential. For all MMMLU experiments, we simply perform a single run with each model to compute the score, without aggregating results from multiple runs.

Furthermore, we evaluate MMMLU using language-specific prompts\footnote{In the original instructions are provided in English. To ensure accurate linguistic evaluation, we use the language-specific prompts. Detailed MMMLU test settings and the modified instructions are provided in Appendix~\ref{appendix:mmmlu_prompts}.} and, to assess the impact of interference when integrating target language data, we include two SFT variants: one trained exclusively on English data (\textit{SFT-eng}) and another trained only on target language data (\textit{SFT-tgt}). We conduct these experiments on five models across three languages.

\begin{table*}[t]
\centering
\resizebox{1.0\textwidth}{!}{%
  \begin{tabular}{lc|cc|cc|cc|cc|cc|cc} 
    \toprule
    \multirow{4}{*}{\textbf{Model}} & \multirow{4}{*}{\textbf{Method}} 
      & \multicolumn{4}{c|}{\textbf{High-Resource}} 
      & \multicolumn{4}{c|}{\textbf{Medium-Resource}} 
      & \multicolumn{4}{c}{\textbf{Low-Resource}} \\
    \cmidrule(lr){3-6} \cmidrule(lr){7-10} \cmidrule(lr){11-14}
      & 
      & \multicolumn{2}{c}{Chinese} & \multicolumn{2}{c|}{German} 
      & \multicolumn{2}{c}{Korean} & \multicolumn{2}{c|}{Indonesian} 
      & \multicolumn{2}{c}{Swahili} & \multicolumn{2}{c}{Yoruba} \\
    \cmidrule(l){3-14}
      & 
      & Target    & English   & Target    & English  
      & Target    & English   & Target    & English  
      & Target    & English   & Target   & English \\
    \midrule
    \multirow{2}{*}{\textbf{Llama-2-7B}}  
      & SFT   & 36.0$\pm$\std{1.60} & 35.6$\pm$\std{1.61} 
              & 31.7$\pm$\std{1.55} & 36.9$\pm$\std{1.61} 
              & 27.4$\pm$\std{1.49} & 37.1$\pm$\std{1.61} 
              & 30.0$\pm$\std{1.53} & 35.8$\pm$\std{1.60} 
              & 23.9$\pm$\std{1.42} & 36.2$\pm$\std{1.60} 
              & 26.0$\pm$\std{1.46} & 36.2$\pm$\std{1.60} \\
      & CLO   & \textbf{36.7}$\pm$\std{1.61} & \textbf{37.1}$\pm$\std{1.60} 
              & \textbf{32.7}$\pm$\std{1.56} & \textbf{38.1}$\pm$\std{1.62} 
              & \textbf{30.1}$\pm$\std{1.53} & \textbf{37.9}$\pm$\std{1.62} 
              & \textbf{31.7}$\pm$\std{1.55} & \textbf{36.6}$\pm$\std{1.61} 
              & \textbf{28.6}$\pm$\std{1.51} & \textbf{37.3}$\pm$\std{1.61} 
              & \textbf{29.0}$\pm$\std{1.51} & \textbf{37.6}$\pm$\std{1.62} \\
    \midrule
    \multirow{2}{*}{\textbf{Llama-2-13B}}  
      & SFT   & 48.2$\pm$\std{1.67} & 58.6$\pm$\std{1.64} 
              & 51.1$\pm$\std{1.67} & 59.6$\pm$\std{1.64} 
              & 37.9$\pm$\std{1.62} & 53.6$\pm$\std{1.66} 
              & 46.7$\pm$\std{1.66} & 60.9$\pm$\std{1.63} 
              & 26.6$\pm$\std{1.47} & 57.4$\pm$\std{1.65} 
              & 25.1$\pm$\std{1.45} & 60.3$\pm$\std{1.63} \\
      & CLO   & \textbf{51.3}$\pm$\std{1.67} & \textbf{58.8}$\pm$\std{1.64} 
              & \textbf{52.3}$\pm$\std{1.67} & \textbf{59.8}$\pm$\std{1.64} 
              & \textbf{38.7}$\pm$\std{1.62} & \textbf{57.6}$\pm$\std{1.65} 
              & \textbf{47.8}$\pm$\std{1.67} & \textbf{61.3}$\pm$\std{1.62} 
              & \textbf{32.3}$\pm$\std{1.56} & \textbf{58.7}$\pm$\std{1.64} 
              & \textbf{27.2}$\pm$\std{1.48} & \textbf{60.7}$\pm$\std{1.63} \\
    \midrule
    \multirow{2}{*}{\textbf{Llama-3-8B}}  
      & SFT   & 69.9$\pm$\std{1.53} & 76.3$\pm$\std{1.42} 
              & 62.9$\pm$\std{1.61} & 73.1$\pm$\std{1.45} 
              & 46.7$\pm$\std{1.66} & 64.9$\pm$\std{1.59} 
              & 57.0$\pm$\std{1.65} & \textbf{75.2}$\pm$\std{1.44} 
              & 42.0$\pm$\std{1.64} & 75.0$\pm$\std{1.44} 
              & 29.6$\pm$\std{1.52} & 76.0$\pm$\std{1.42} \\
      & CLO   & \textbf{70.6}$\pm$\std{1.52} & \textbf{77.1}$\pm$\std{1.40} 
              & \textbf{64.1}$\pm$\std{1.60} & \textbf{74.7}$\pm$\std{1.48} 
              & \textbf{57.7}$\pm$\std{1.65} & \textbf{73.4}$\pm$\std{1.47} 
              & \textbf{58.7}$\pm$\std{1.64} & 75.0$\pm$\std{1.44} 
              & \textbf{42.6}$\pm$\std{1.65} & \textbf{75.9}$\pm$\std{1.43} 
              & \textbf{29.8}$\pm$\std{1.53} & \textbf{76.3}$\pm$\std{1.42} \\
    \midrule
    \multirow{2}{*}{\textbf{Mistral-7B-v0.1}}  
      & SFT   & 56.3$\pm$\std{1.63} & 70.6$\pm$\std{1.52} 
              & 47.6$\pm$\std{1.67} & 67.9$\pm$\std{1.56} 
              & 29.2$\pm$\std{1.52} & \textbf{70.6}$\pm$\std{1.52} 
              & 51.1$\pm$\std{1.67} & 72.3$\pm$\std{1.49} 
              & 34.3$\pm$\std{1.58} & \textbf{72.3}$\pm$\std{1.49} 
              & 31.7$\pm$\std{1.55} & 68.2$\pm$\std{1.55} \\
      & CLO   & \textbf{61.1}$\pm$\std{1.65} & \textbf{70.7}$\pm$\std{1.52} 
              & \textbf{53.0}$\pm$\std{1.66} & \textbf{72.3}$\pm$\std{1.49} 
              & \textbf{48.0}$\pm$\std{1.67} & 58.8$\pm$\std{1.64} 
              & \textbf{53.4}$\pm$\std{1.66} & \textbf{75.7}$\pm$\std{1.43} 
              & \textbf{38.8}$\pm$\std{1.63} & 69.8$\pm$\std{1.53} 
              & \textbf{31.9}$\pm$\std{1.55} & \textbf{74.7}$\pm$\std{1.45} \\
    \midrule
    \multirow{2}{*}{\textbf{Qwen2.5-3B}}  
      & SFT   & 83.1$\pm$\std{1.25} & 81.6$\pm$\std{1.29} 
              & 70.3$\pm$\std{1.52} & 81.3$\pm$\std{1.30} 
              & 67.7$\pm$\std{1.56} & 80.9$\pm$\std{1.31} 
              & 65.3$\pm$\std{1.59} & 81.8$\pm$\std{1.29} 
              & 36.6$\pm$\std{1.61} & 82.9$\pm$\std{1.25} 
              & 27.0$\pm$\std{1.48} & 80.9$\pm$\std{1.29} \\
      & CLO   & \textbf{83.4}$\pm$\std{1.24} & \textbf{82.3}$\pm$\std{1.27} 
              & \textbf{73.1}$\pm$\std{1.48} & \textbf{82.0}$\pm$\std{1.28} 
              & \textbf{68.8}$\pm$\std{1.55} & \textbf{82.0}$\pm$\std{1.28} 
              & \textbf{67.1}$\pm$\std{1.57} & \textbf{83.1}$\pm$\std{1.25} 
              & \textbf{39.8}$\pm$\std{1.63} & \textbf{83.1}$\pm$\std{1.26} 
              & \textbf{29.6}$\pm$\std{1.52} & \textbf{81.7}$\pm$\std{1.31} \\
    \bottomrule
  \end{tabular}
}
\caption{Zero-shot evaluation results for BELEBELE. In each cross-lingual setting, “Target” refers to accuracy for the target language, while “English” indicates accuracy for English. The standard deviations ($\pm$) are also reported.}
\label{tab:belebele}
\end{table*}

\section{Experimental Results}
We present the experimental results on three key benchmarks: AlpacaEval, which measures instruction following ability, BELEBLE, which evaluate the models' MRC abilities, and MMMLU, which assesses the reasoning performance.

\paragraph{Instruction Following Ability}
We evaluate the instruction following ability of the models using AlpacaEval, and the results are presented in Table~\ref{tab:alpaca}.\footnote{Following~\citet{marchisio2024understanding}, we report additional results for the target language evaluation in Table~\ref{tab:zero-shot}, wherein the instruction ‘Please answer in the same language as the input’ is translated into the target language and appended to the original prompt.} The SFT+DPO enhances the performance of the SFT baseline in target languages, maintaining comparable performance in English for high-resource languages. However, this configuration yields only modest improvements in medium-resource languages and demonstrates a tendency to prioritize English output in low-resource languages. Conversely, the CLO consistently surpasses the SFT across all base models and languages, achieving a win rate exceeding 50\% in all cases. CLO demonstrates substantial advancements in medium-resource languages, indicative of its consistent alignment with the target language. Even though CLO outperforms SFT in English, SFT+DPO only partially enhances SFT and fails to close the performance gap with CLO. This discrepancy is particularly pronounced in low-resource languages, where CLO significantly outperforms SFT+DPO, a result possibly attributable to the detrimental effect of excessive parameter fluctuations during fine-tuning due to insufficient embedded language knowledge.

Remarkably, the Llama-3, with its extensive internal knowledge, and Qwen-2.5, known for its outstanding multilingual capabilities, exhibit the greatest performance gains when using CLO, highlighting CLO's efficacy in leveraging internal English knowledge for enhanced cross-lingual alignment with the target language. These findings suggest that SFT training may overly prioritize English, with target language data inducing disruptive parameter fluctuations that further undermine the model’s multilingual capabilities. In contrast, CLO offers a more equitable adaptation approach, sustaining robust English performance while adeptly accommodating the target language. Moreover, the results evaluated using the original AlpacaEval prompt, without our language-specific prompts, are presented in Appendix~\ref{appendix:alpaca_prompt}, with outcomes generally showing similar performance in the target language. Upon manual review of the differently evaluated results, we observed discrepancies only when responses were provided in another language at the word level.\footnote{Excluding the Mistral, no word-level confusion is observed; hence, we do not provide additional Word-level Confusion analysis~\cite{marchisio2024understanding}.} These observations suggest that our language-specific prompt is adequately assessing the model's capability in handling the target language.

\paragraph{MRC Performances}
The results are reported in Table~\ref{tab:belebele}. Overall, CLO exhibits superior performance compared to SFT by effectively adapting target language. By contrast, SFT either struggles to adapt fully to the target languages or remains overly reliant on English, resulting in decreased accuracy. 

Notably, Qwen2.5, despite having the smallest parameter scale among the compared models, demonstrates particularly strong performance in Chinese. We believe that this is because Qwen2.5 was pre-trained on a substantial amount of Chinese data, facilitating its adaptation to Chinese-specific tasks. Furthermore, Qwen2.5 also achieves high accuracy in the other target languages, suggesting that its multilingual pre-training facilitates robust cross-lingual adaptation~\cite{yang2024qwen2}.

\paragraph{Reasoning Performance}
We evaluate the models on MMMLU, with the results summarized in Table~\ref{tab:mmmlu}.\footnote{Most models experience extraction failures in less than 0.8\% of cases, indicating that such failures have minimal impact on performance evaluation.} The proposed CLO generally demonstrates higher performance across most of the languages and models tested. The CLO Llama-3-8B models exhibits outstanding performance, achieving higher scores compared to the SFT. This underscores the significant enhancements gained through leveraging the model's extensive internal knowledge via the CLO method. In contrast, the SFT tends to show relatively high performance in English but lower performance in target languages, especially in low-resource settings. This suggests that traditional SFT methods may not adapt well when limited data is available and tend to learn in an English-centric manner. While applying DPO to SFT models using cross-lingual datasets can generally improve performance, we observe that this approach is cost-inefficient and does not consistently guarantee enhanced performance in target languages. In some cases, SFT+DPO still exhibit English-centric learning patterns.

\begin{table*}[t]
\centering
\resizebox{0.6\textwidth}{!}{%
\begin{tabular}{ll|cc|cc|cc}
\toprule
\multicolumn{1}{c}{\multirow{2}{*}{\textbf{Model}}} &
\multicolumn{1}{c|}{\multirow{2}{*}{\shortstack{\textbf{Method}}}} &
\multicolumn{2}{c|}{\textbf{Chinese}} &
\multicolumn{2}{c|}{\textbf{Korean}} &
\multicolumn{2}{c}{\textbf{Swahili}} \\
\cmidrule(l){3-8}
\multicolumn{1}{c}{} &
\multicolumn{1}{c|}{} &
\textbf{Target} & \textbf{English} &
\textbf{Target} & \textbf{English} &
\textbf{Target} & \textbf{English} \\
\midrule
\multirow{5}{*}{\textbf{Llama-2-7B}} &
SFT-eng   & --     & 29.40  & --     & 29.40  & --     & 29.40 \\
& SFT-tgt & 27.19  & --     & 25.31  & --     & 22.46  & --    \\
\cline{2-8}
& SFT     & 27.12  & 29.78  & 23.47  & 26.01  & 17.21  & 28.99 \\
& SFT+DPO & 26.59  & 31.00  & 28.00  & 29.88  & 19.96  & 28.10 \\
& CLO     & \textbf{28.11}  & \textbf{31.24}  & \textbf{29.09}  & \textbf{31.54}  & \textbf{24.09}  & \textbf{30.14} \\
\midrule
\multirow{5}{*}{\textbf{Llama-2-13B}} &
SFT-eng   & --     & 41.81  & --     & 41.81  & --     & 41.81 \\
& SFT-tgt & 31.51  & --     & 36.80  & --     & 22.56  & --    \\
\cline{2-8}
& SFT     & 31.17  & 42.47  & 34.39  & 37.69  & 21.38  & 36.77 \\
& SFT+DPO & 33.26  & 43.75  & 26.79  & 40.22  & 19.83  & \textbf{46.32} \\
& CLO     & \textbf{34.17}  & \textbf{47.20}  & \textbf{39.70}  & \textbf{44.14}  & \textbf{26.78}  & 41.68 \\
\midrule
\multirow{5}{*}{\textbf{Llama-3-8B}} &
SFT-eng   & --     & 49.13  & --     & 49.13  & --     & 49.13 \\
& SFT-tgt & 38.55  & --     & 29.61  & --     & 29.05  & --    \\
\cline{2-8}
& SFT     & 39.36  & 53.00  & 25.31  & 50.61  & 27.59  & 44.48 \\
& SFT+DPO & 40.91  & 56.36  & 27.48  & 50.71  & 28.86  & \textbf{55.97} \\
& CLO     & \textbf{41.99}  & \textbf{57.55}  & \textbf{32.73}  & \textbf{55.57}  & \textbf{33.38}  & 55.82 \\
\midrule
\multirow{5}{*}{\textbf{Mistral-7B}} &
SFT-eng   & --     & 50.67  & --     & 50.67  & --     & 50.67 \\
& SFT-tgt & 33.16  & --     & 27.65  & --     & 28.61  & --    \\
\cline{2-8}
& SFT     & 33.74  & 52.11  & 25.94  & 34.11  & 26.68  & 49.95 \\
& SFT+DPO & \textbf{37.02}  & 52.03  & 26.77  & 36.73  & \textbf{28.42}  & \textbf{53.39} \\
& CLO     & 34.05  & \textbf{58.74}  & \textbf{28.31}  & \textbf{50.95}  & 28.07  & 50.73 \\
\midrule
\multirow{2}{*}{\textbf{Qwen2.5-3B}} &
SFT     & 46.34  & 55.70  & 35.90  & 56.01  & 26.22  & 57.19 \\
& CLO     & \textbf{52.10}  & \textbf{60.52}  & \textbf{41.94}  & \textbf{61.49}  & \textbf{29.80}  & \textbf{60.92} \\
\bottomrule
\end{tabular}
}%
 \caption{MMMLU evaluation results. The performance of \textit{CLO}, \textit{SFT}, \textit{SFT+DPO}, \textit{SFT-eng}, and \textit{SFT-tgt} models trained on CLO datasets in Chinese, Korean, and Swahili across four base models. \textit{SFT-eng} indicates the performance of models where SFT is performed using only English data, while \textit{SFT-tgt} denotes the performance of models where SFT is performed using only target language data.}
\label{tab:mmmlu}
\end{table*}

We observed that, except for the Mistral, the performances of the \textit{SFT-eng} model are similar to that of the SFT models trained on both languages. These results suggest that in high-resource languages like Chinese, there is no loss in English performance when additional language data is included. The inclusion of the Chinese dataset appears to act as data augmentation, enhancing the model's robustness and resulting in better performance compared to training on English data alone. In contrast, for medium-resource languages like Korean (excluding the Llama-3), \textit{SFT-eng} outperforms the SFT model trained on both English and Korean. This discrepancy becomes even more pronounced in the low-resource setting of Swahili. These findings indicate that incorporating medium-resource Korean and low-resource Swahili data can potentially degrade the model's English capabilities. We speculate that in the case of the Llama-3, Korean is also learned with relatively high-resource data, which might explain why training on both languages yields better performance compared to the traditional SFT approach. This hypothesis is further supported by the performance of the \textit{SFT-tgt}. In the high-resource Chinese setting, the \textit{SFT-tgt} model's performance is comparable to that of the SFT model trained on both datasets. However, in medium-resource Korean and low-resource Swahili, the \textit{SFT-tgt} model records higher performance than the SFT, providing clearer evidence to support our speculation.

Additionally, since Llama-3-8b exhibited the most pronounced changes in previous experiments, we conducted a category-wise analysis of its MMMLU performance in the target languages, as presented in Appendix~\ref{Appendix:MMMLU_analysis}. In conclusion, the CLO Llama-3 demonstrated strong performance across a wide range of categories, whereas the SFT model showed decreased performance in specialized domains, highlighting the limitations of the SFT method in low-data scenarios.

\begin{figure*}[t]
 \centering
\includegraphics[width=1.0\linewidth]
{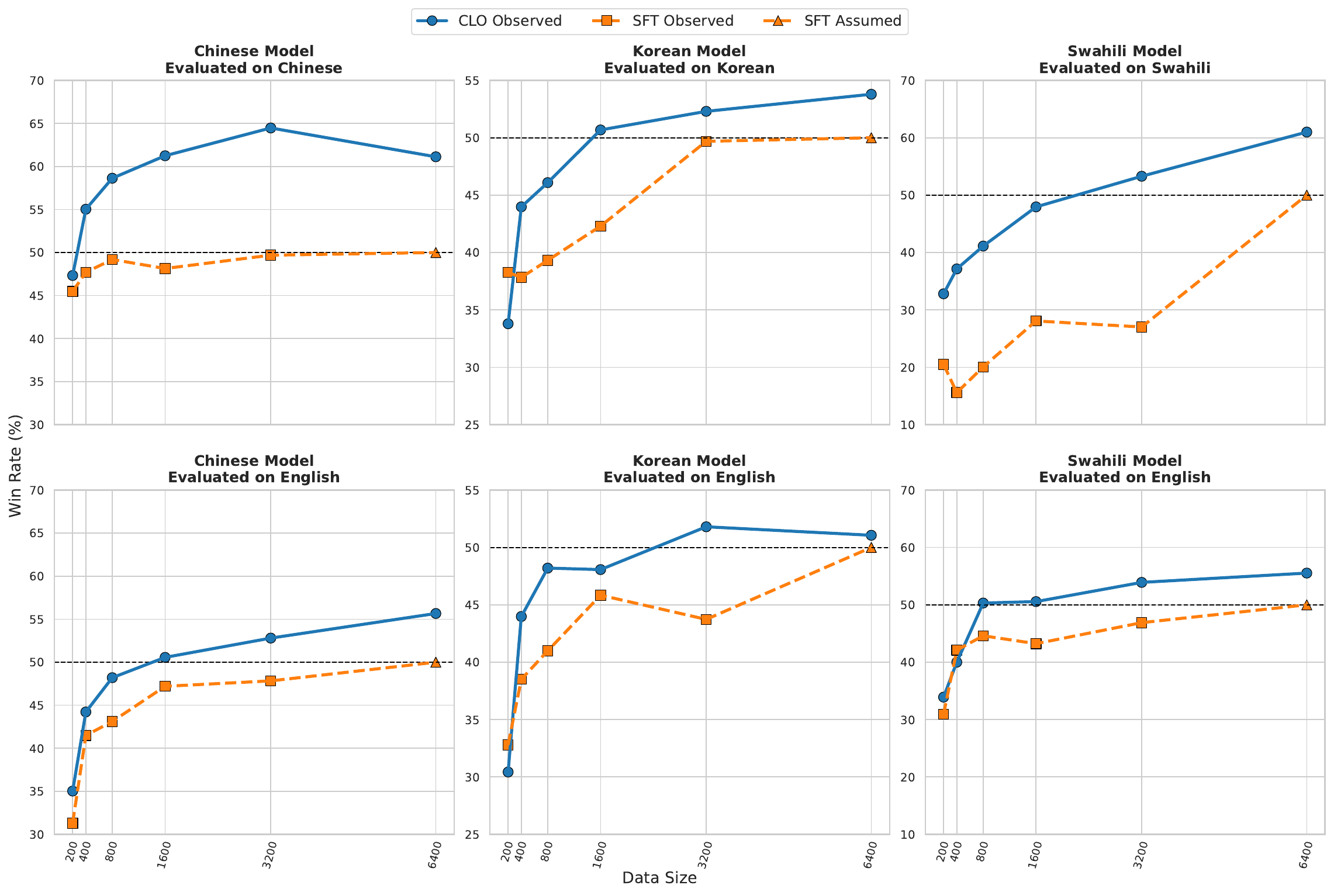}
\caption{Comparison of win rates between CLO and SFT on Llama-2-7B models trained with varying amounts of data, evaluated against a SFT with 6,400 pair examples on the AlpacaEval. The 'SFT Assumed' baseline is assigned a win rate of 50\%, as it compares identical models.}
 \label{fig:dataperformance_main}
\end{figure*}

\section{Effect of Training Data Size}
\label{sec:7}

To evaluate the impact of training data size on the performance of CLO and SFT, we conduct experiments by varying the amount of training data. Specifically, we adjust the number of training examples per language to 200, 400, 800, 1600, 3200, and 6400 based on single-turn English data. For comparison, we utilize the AlpacaEval test setup employed, and each model trained with a different data size is compared against the SFT model trained on 6,400 pair examples. The experimental results of Llama-2-7b are depicted in Figure~\ref{fig:dataperformance_main} and the results for Llama-3-8b are presented in Appendix~\ref{appendix:Llama3-data}.

The results of Llama-2 reveal that both CLO and SFT achieve relatively rapid performance improvements with smaller amounts of data in Chinese, a high-resource language. However, for SFT, performance improvements in Korean, a medium-resource language, are observable only when the training data exceeds 3,200 pairs. Moreover, in Swahili, a low-resource language, there is a significant performance difference between models trained with 3200 and 6,400 pairs. In contrast, CLO demonstrates efficient performance enhancement across all languages, even in low-data environments. Notably, in Swahili, CLO exhibits exceptional performance by achieving results comparable to the SFT model trained on 6,400 pairs using only 1600 pairs.

We find that in Llama-3, the SFT method achieves similar performance with only 1,600 pairs as it does with 6,400 pairs for high-resource language Chinese and medium-resource language Korean. However, for the low-resource language Swahili, the SFT method requires 6,400 pairs to achieve comparable performance. On the other hand, the CLO method attains performance comparable to the SFT model trained on 6,400 pairs using only 400 pairs in the target language. Both the results of Llama-2 and Llama-3 indicate that transfer to low-resource languages is more challenging for the SFT than for the CLO.

Overall, our findings confirm that CLO attains faster performance improvements compared to SFT in Chinese, Korean, and Swahili. This demonstrates that CLO can facilitate rapid language transfer even with relatively small amounts of data. Particularly in Swahili, CLO achieves performance similar to the SFT model trained on 6,400 pairs using merely 1,600 pairs. In summary, CLO enhances performance more efficiently than SFT across diverse language environments. Especially in low-resource languages, CLO outperforms the existing SFT models with substantially less training data. Conversely, SFT exhibits a strong dependence on the quantity of training data in medium and low-resource languages, indicating potential limitations in performance improvement when sample data collection is infeasible.

\section{Ablation Studies}
\subsection{Comparison with Full Tuning}

\begin{table}[t]
\centering
\resizebox{0.98\linewidth}{!}{
\begin{tabular}{lcccccc}
\toprule
\multirow{2}{*}{\textbf{Models}} & \multicolumn{2}{c}{\textbf{Target Language}} & \multicolumn{2}{c}{\textbf{English Evaluation}} \\
& \textit{Win (\%)} & \textit{Lose (\%)} & \textit{Win (\%)} & \textit{Lose (\%)}\\
\midrule
\textbf{Chinese} & \textbf{50.68} & 49.07  & \textbf{54.41} & 45.59 \\
\textbf{Korean} & \textbf{52.73} & 46.40  & \textbf{52.67} & 47.33 \\
\textbf{Swahili} & 29.69 & \textbf{69.57} & \textbf{50.06} & 49.94 \\
\bottomrule
\end{tabular}
}
\caption{Comparison of AlpacaEval generation performance between the only trained attention Llama-2 CLO ($\pi_{\theta_{\text{attn}}}$) and the all parameter trained Llama-2 CLO ($\pi_{\theta}$) models trained on Chinese, Korean, and Swahili.}
\label{tab:ablation}
\end{table}

According to \citet{anonymous2024neuronlevel}, certain types of knowledge related to language can be mostly stored in the attention layers of language models, and important neurons are concentrated in deeper layers. Based on these results, we applied a method in CLO where only the attention layers are trained during parameter updates, and compared the performance with updating all parameters, as shown in Table~\ref{tab:ablation}.

Experimental results confirmed that updating only the attention layers maintains or even improves the model's performance. This indicates that in our method, language alignment for knowledge representation and utilization can be achieved by training only the attention layers, suggesting the potential of an efficient model update strategy. However, our results found that in Swahili, the attention-only training CLO method showed a significant performance drop compared to full training. This result suggests that aligning the language by training only the attention layers in low-resource languages like Swahili is challenging.

\begin{table}[t]
\centering
\resizebox{0.8\linewidth}{!}{
\begin{tabular}{lcc}
\toprule
\textbf{NLL Loss} & \textbf{Swahili} & \textbf{English} \\
\midrule
Target-Only (ours)    & 83.0             & 65.4             \\
Target \& English        & 74.5             & 67.8             \\
\bottomrule
\end{tabular}
}
\caption{Comparison of win rates on the AlpacaEval dataset for SFT and CLO variants in Swahili, both trained with 6,400 pair examples on the Llama-3-8B model. Results are presented using only target language NLL loss and both target and English NLL losses.}
\label{tab:nll_study}
\end{table}

\subsection{NLL Loss Analysis}
Furthermore, we conducted ablation experiments to assess the effect of using a combined target and English NLL loss versus employing only the target language NLL loss, with the results presented in Table~\ref{tab:nll_study}. The findings indicate that incorporating both losses inadvertently biases the model toward English responses, as evidenced by the higher performance on English at the expense of the target language. Consequently, we adopt the approach that considers only the target language NLL loss, which better maintains a balanced performance between the target and English languages.

\section{Conclusion}
In this paper, we introduced \textbf{Cross-Lingual Optimization (CLO)}, an effective strategy for transferring English-centric LLMs to target languages while preserving their English capabilities. Leveraging publicly available English SFT data and translation models, CLO facilitates cross-lingual transfer without the need for extensive target language data.

We conduct experiments using five LLMs across six languages with varying resource levels: two high-resource (Chinese, German), two medium-resource (Korean, Indonesian), and two low-resource (Swahili, Yoruba). Our results show that CLO outperforms SFT in both acquiring target language proficiency and maintaining English performance. Notably, in the low-resource language, CLO achieved superior results with only 3,200 pairs, surpassing SFT models trained on twice the amount of data. We found that traditional SFT is particularly sensitive to data quantity in medium and low-resource languages, often leading the model to either overly rely on its English knowledge or diminish it when data is scarce, resulting in insufficient adaptation to the target language. In contrast, CLO's approach for responding in the target language enables it to utilize embedded language knowledge more effectively, leading to better performance even with less data.

\section*{Limitations}
\paragraph{Multilinguality} This study focuses on the cross-lingual transfer of an English-centric large language model to a specific target language rather than expanding to multiple languages simultaneously. Consequently, our research does not address the enhancement of multilingual performance across several languages at once. This limitation suggests the need for future work to explore methods for transferring to multiple languages concurrently to improve overall multilingual capabilities.

\paragraph{Training Data} Our approach relies on translated data to address the scarcity of human-produced target-language instruction data, particularly for low-resource languages. Translation models can introduce semantic distortions or uncertainties, and our current work does not explicitly quantify how these factors might affect the model’s performance. However, CLO demands the use of datasets that are aligned between English and the target language, which is a resource that is extremely costly and often impractical to curate manually at scale. Given these constraints, employing a translation model becomes a necessity, especially for low-resource languages.

Since both our baseline SFT and the proposed CLO method operate on the same translated training set, any translation artifacts or alignment issues are likely to impact both approaches uniformly, thus ensuring a fair comparison of their relative performance. Moreover, manual inspection suggests that major translation errors were minimal, and our empirical results indicate that CLO consistently outperforms SFT across language settings, suggesting that minor translation imperfections do not undermine the advantages of leveraging embedded English knowledge to enhance target-language capability.

\paragraph{Limited Scope of Languages} Our study was limited to experiments involving English and six target languages (Chinese, German, Korean, Indonesian, Swahili, and Yoruba). We selected these six because extending the analysis to additional languages would require training and evaluating three separate models (SFT, SFT+DPO, and CLO), which would introduce significant additional costs and time constraints. Consequently, there are limitations in generalizing the results to all languages, and it is necessary to examine the potential for extending our methods to a wider variety of languages in future work.

\paragraph{Evaluation on Language-Specific Data} Our experiments evaluated language models using AlpacaEval datasets translated by GPT-4o and the MMMLU dataset, which was translated into respective languages by professional human translators, correcting the initial misstatement. Additionally, we assessed Machine Reading Comprehension abilities using the BELEBELE~\cite{bandarkar2023belebele} dataset, constructed and reviewed by human translators. While these datasets are suitable for measuring general model performance, they do not fully capture the model's ability to respond appropriately to data involving linguistic characteristics or cultural contexts specific to each language. Consequently, we were unable to thoroughly evaluate the model’s handling of language-specific nuances or culturally relevant content.

\paragraph{Applicability to Other Methods} The proposed CLO method was experimented with only by applying it to the DPO Direct Preference Optimization methodology, although it can potentially be applied to various preference optimization algorithms~\cite{ethayarajh2024kto, hong2024orpo, xie2024exploratory}. However, since our experiments did not study the generality of CLO across these different preference optimization methods, further research is needed to verify whether CLO can guarantee performance improvements in other methodologies.

\paragraph{Computational Efficiency}
The CLO method performs parameter updates based on a reference model and trains only a subset of the total parameters, resulting in a lower computational cost than DPO while incurring a slightly additional cost relative to SFT training. In our environment, the GPU memory allocation required for CLO is up to 55\% higher than that of conventional SFT training, whereas it is approximately 30\% percent lower than that required by DPO. Moreover, our method does not allow for an exact computation of FLOPs, making precise inference of the training cost difficult. In our experiments, the training time for CLO is nearly identical to or slightly higher than that of SFT.


\bibliography{custom}

\appendix
\onecolumn

\section{Details of Cross-Lingual Loss}
\label{appendix:cross-lingual-loss}

\paragraph{Cross-Lingual Dataset Preparation}
We generate target language prompts \(\mathbf{x}_{\ell}\) and target language responses \(\mathbf{y}_{\ell}\) by translating the English prompts \(\mathbf{x}_{\text{en}}\) and English responses \(\mathbf{y}_{\text{en}}\) in the existing English preference dataset using a translation model.

For \(\mathbf{x}_{\text{en}}\), we map the \emph{chosen response} to \(\mathbf{y}_{\text{en}}\) and the \emph{rejected response} to \(\mathbf{y}_{\ell}\), constructing the English training data. This is to prevent the model from responding in the target language when given an English prompt, thereby preserving the existing English knowledge within the model.

Conversely, for \(\mathbf{x}_{\ell}\), we map the \emph{chosen response} to \(\mathbf{y}_{\ell}\) and the \emph{rejected response} to \(\mathbf{y}_{\text{en}}\), constructing the target language data. This is to suppress the model's tendency to respond in English when given a target language instruction, encouraging it instead to utilize its English knowledge to generate outputs in the target language.

\paragraph{Cross-Lingual Loss Function}

Our proposed CLO introduces a new loss function that differs from the standard DPO loss by utilizing cross-lingual data pairs within the \emph{same batch} to explicitly teach the model the correspondence between input and output languages.

The overall objective function of CLO is defined as:
\begin{equation}
\begin{aligned}
\mathcal{L}_{\text{CLO}} = \lambda \cdot \mathcal{L}_{\text{SFT}} + (1-\lambda) \cdot \mathcal{L}_{\text{CL}}
\end{aligned}
\end{equation}

Here, \(\mathcal{L}_{\text{SFT}}\) is the supervised fine-tuning loss that promotes the language model to generate correct outputs for the target language only. It is calculated using the Negative Log-Likelihood (NLL) over the target-language data in the batch:
\begin{equation}
\begin{aligned}
\mathcal{L}_{\text{SFT}} = \frac{1}{N} \sum_{i=1}^{N} \left[
 - \log \pi_{\theta_{\text{att}}}(y_{\ell}^{(i)} \mid x_{\ell}^{(i)}) 
\right],
\end{aligned}
\end{equation}

where \(N\) is the batch size, and \(\bigl(x_{\ell}^{(i)}, y_{\ell}^{(i)}\bigr)\) are the target language input-output pairs in the batch.

\(\mathcal{L}_{\text{CL}}\) is our proposed cross-lingual loss that encourages the model to generate outputs in the correct language based on the input language by utilizing cross-lingual data pairs within the same batch. It is defined as:
\begin{equation}
\begin{aligned}
\mathcal{L}_{\text{CL}} = -\frac{1}{N} \sum_{i=1}^{N} \Big[ & \log \sigma \left( z_{\text{en}}^{(i)} \right) \\
& + \log \sigma \left( z_{\ell}^{(i)} \right) \Big],
\end{aligned}
\end{equation}
where \( z_{\text{en}}^{(i)} \) and \( z_{\ell}^{(i)} \) are defined respectively as:
\begin{equation}
\begin{aligned}
z_{\text{en}}^{(i)} = \beta \left( \log \frac{\pi_{\theta_{\text{att}}}(y_{\text{en}}^{(i)} \mid x_{\text{en}}^{(i)})}{\pi_{\text{ref}}(y_{\text{en}}^{(i)} \mid x_{\text{en}}^{(i)})} - \log \frac{\pi_{\theta_{\text{att}}}(y_{\ell}^{(i)} \mid x_{\text{en}}^{(i)})}{\pi_{\text{ref}}(y_{\ell}^{(i)} \mid x_{\text{en}}^{(i)})} \right),
\end{aligned}
\end{equation}
\begin{equation}
\begin{aligned}
z_{\ell}^{(i)} = \beta \left( \log \frac{\pi_{\theta_{\text{att}}}(y_{\ell}^{(i)} \mid x_{\ell}^{(i)})}{\pi_{\text{ref}}(y_{\ell}^{(i)} \mid x_{\ell}^{(i)})} - \log \frac{\pi_{\theta_{\text{att}}}(y_{\text{en}}^{(i)} \mid x_{\ell}^{(i)})}{\pi_{\text{ref}}(y_{\text{en}}^{(i)} \mid x_{\ell}^{(i)})} \right).
\end{aligned}
\end{equation}
In these equations, \(\pi_{\theta}\) denotes the parameterized policy, \(\pi_{\text{ref}}\) represents the reference policy, \(\beta\) is a hyperparameter indicating the strength of the KL constraint, \(\sigma\) is the sigmoid function, and the superscript \((i)\) refers to the \(i\)-th sample in the batch.

By incorporating both the preferred and rejected responses in different languages within the same batch, the model is explicitly guided to increase the likelihood of outputs in the appropriate language for a given input language while decreasing the likelihood of outputs in the incorrect language. This mechanism ensures that the model not only learns the correspondence between input and output languages but also effectively utilizes its English knowledge for the target language without losing its proficiency in English.

\paragraph{Comparison with Standard DPO on Cross-Lingual Data}

It's important to emphasize the key differences between our proposed CLO and simply applying the standard DPO to cross-lingual augmented data. In the standard DPO approach with cross-lingual data augmentation, the loss function is applied independently to each language's data, and the model does not utilize cross-lingual data pairs within the same batch. That is, the model may only learn the correspondence between input and output languages \emph{implicitly} and might not effectively utilize the relationships between accepted and rejected responses across languages.

In contrast, our CLO method constructs the loss function by leveraging English and target language data pairs within the \emph{same batch}, as shown in Equations (9)–(12). By pairing each English input-output pair with its corresponding translated target language input-output pair within the batch, the model is explicitly taught to respond in English when given English input and in the target language when given target language input. Moreover, by contrasting the probabilities of the accepted and rejected responses across languages within each sample in the batch, the model prevents knowledge loss and encourages the utilization of English knowledge in the target language.

Through this approach, CLO enables the model to select the correct output language according to the input language and allows for effective target language transfer using its English knowledge. This is fundamentally different from just applying DPO on cross-lingual data augmentation, where the model might not sufficiently learn to adjust the output language based on the input language, potentially leading to suboptimal cross-lingual transfer and loss of English proficiency.

By combining cross-lingual data augmentation with our newly designed batch-based loss function, CLO ensures that the model preserves its English knowledge while effectively transferring it to the target language, achieving superior performance compared to methods that only use data augmentation with standard DPO.

\section{Hyperparameters}
\label{appendix:hyperparameters}
In our experiments, we utilized a fixed setup with a server equipped with 8 NVIDIA A100 GPUs, each with 80GB of memory. The training hyperparameters were set as follows:

\begin{itemize}
    \item \textbf{Trade-off Parameter} ($\lambda$): \texttt {0.5}
    \item \textbf{Learning rate}: \texttt{5e-5}
    \item \textbf{Minimum learning rate}: \texttt{1.1e-6}
    \item \textbf{Max sequence length}: \texttt{3000}
    \item \textbf{Beta} ($\beta$): \texttt{0.1}
    \item \textbf{Training batch size}: \texttt{8}
\end{itemize}

All models used in the experiments (except for the SFT + DPO in Table~\ref{tab:mmmlu}) were trained for only 1 epoch. During training, the model with the lowest validation loss was selected for use. The SFT + DPO models experienced an additional training process using the same cross-lingual dataset as CLO during the DPO phase for each SFT model.

For text generation, we used the following fixed generation configurations:

\begin{itemize}
    \item \textbf{Top-p (nucleus sampling)}: \texttt{0.9}
    \item \textbf{Temperature}: \texttt{0.6}
\end{itemize}

We set the trade-off parameter $\lambda=0.5$ for all models and languages in our experiments. While tuning $\lambda$ for each model and language individually may yield further performance improvements, we use a fixed value of $0.5$ in order to report consistent results across different models and languages and to ensure fair comparison.

\section{AlpacaEval Setup}
\label{appen:alpacaeval}

We detail the prompts used for evaluating and ranking LLMs. To more accurately measure the multilingual capabilities of the models, except for English, we modified the existing prompts for the other languages (Chinese, Korean, and Swahili). Specifically, we included additional instructions to evaluate responses that are in a different language from the instruction. If a model responds in a language different from the one used in the instruction, this is reflected in its evaluation. This adjustment allows us to assess the models' performance in multilingual settings more effectively. Conversely, when evaluating English performance, we used the original AlpacaEval~\cite{alpaca_eval} prompts without modification. 

The following is the prompt provided to the \texttt{GPT-4o} for evaluation:

\begin{quote}
\textbf{System Prompt:}

\begin{lstlisting}[basicstyle=\ttfamily\footnotesize, breaklines=true]
You are a highly efficient assistant, who evaluates and rank large language models (LLMs) based on the quality of their responses to given prompts. This process will create a leaderboard reflecting the most accurate and human-preferred answers.
\end{lstlisting}

\textbf{User Prompt:}
\begin{lstlisting}[basicstyle=\ttfamily\footnotesize, breaklines=true]
I require a leaderboard for various large language models. I'll provide you with prompts given to these models and their corresponding responses. Your task is to assess these responses, ranking the models in order of preference from a human perspective. Once ranked, please output the results in a structured JSON format for the make_partial_leaderboard function.

## Prompt
{
    "instruction": """{instruction}""",
}

## Model Outputs

Here are the unordered outputs from the models. Each output is associated with a specific model, identified by a unique model identifier.
{
    {
        "model": "m",
        "output": """{output_1}"""
    },
    {
        "model": "M",
        "output": """{output_2}"""
    }
}

## Task
Evaluate and rank the models based on the quality and relevance of their outputs. The ranking should be such that the model with the highest quality output is ranked first. Additionally, since the purpose is to measure multilinguality, if a model responds in a language different from the instruction's language, this should be reflected in the evaluation.
\end{lstlisting}
\end{quote}

The key modification in this prompt is the inclusion of a specific instruction to consider the language of the model's response relative to the instruction's language. By doing so, we aim to evaluate the models' multilinguality more clearly. If a model provides an answer in a language different from the one used in the instruction, this difference is factored into its evaluation, potentially affecting its ranking on the leaderboard.

By employing OpenAI's \texttt{GPT-4o} model for the evaluation, we utilize its advanced understanding and reasoning capabilities to perform a nuanced assessment of the LLMs' responses.

\section{MMMLU Evaluation Setup}
\label{appendix:mmmlu_prompts}
In the original MMMLU test set, all fixed prompts requiring answers are written in English. We found that consistently using English prompts without employing language-specific prompts corresponding to the test set resulted in high standard deviations in performance. Therefore, to more accurately measure adaptability and performance in the target language, we adjusted the prompts to match the input language. The specific prompts for each language are presented in Table~\ref{table:mmmlu_prompts}. Each prompt means exactly same instructions, requiring the answers to multiple-choice questions to be written in a specific format by selecting one of A, B, C, or D. The dataset comprises approximately 14,000 data points.

If the correct answer cannot be extracted from the model's response, it is considered incorrect, as this indicates that the model has not properly followed the instructions. (Since MMMLU's performance focuses on evaluating the model's knowledge, the answer is considered correct as long as the correct answer is inferred, even if the given query is not in English and the response is in English.)

We were concerned that, due to limitations in the capabilities of existing pre-trained models, significant performance disparities between models could arise if the correct answer was inferred but could not be extracted from the response. Since extraction might fail due to issues with regular expression matching, we aimed to enhance the reliability of the extraction process. To this end, the evaluation model included 10 English and 10 target language test samples per target language in its training. The 10 test samples used for training were excluded from the MMMLU performance evaluation. If the answer could not be extracted from the multiple-choice responses, we allowed up to three regeneration attempts. If the answer still could not be extracted after these attempts, the response was considered incorrect.

\begin{table}[ht]
\centering
\resizebox{0.9\linewidth}{!}{
\begin{tabular}{|l|p{14cm}|}
\hline
\textbf{Language} & \textbf{Prompt} \\ \hline
English & Answer the following multiple choice question. Ensure the last line of your response is in the format: 'Answer: \$LETTER' (without quotes) where LETTER is one of ABCD. For example, 'Answer: A'. \\ \hline
Korean & 다음 선택형 질문에 답하십시오. 당신의 응답 마지막 줄을 '답변: \$LETTER' (따옴표 제외) 형식으로 작성하십시오. 여기서 LETTER는 ABCD 중 하나입니다. 예를 들어, '답변: A'로 대답하세요. \\ \hline
Chinese & 回答以下选择题。确保您的回答的最后一行格式为：'答案: \$LETTER'（不带引号），其中 LETTER 是 ABCD 之一。例如，'答案: A'。 \\ \hline
Swahili & Jibu swali lifuatalo la chaguo nyingi. Hakikisha mstari wa mwisho wa jibu lako uko katika muundo: 'Answer: \$LETTER' au 'Jibu: \$LETTER' (bila nukuu) ambapo LETTER ni moja ya ABCD. Kwa mfano, 'Answer: A' au 'Jibu: A'. \\ \hline
\end{tabular}
}
\caption{Language-specific prompts for MMMLU test set}
\label{table:mmmlu_prompts}
\end{table}

\section{Machine Reading Comprehension Performances}
\label{appendix:belebele}
To evaluate the Machine Reading Comprehension (MRC) capabilities of SFT and CLO, we adopt the BELEBELE dataset~\cite{bandarkar2023belebele}. In this benchmark, a question and a passage are provided along with four answer choices. The model’s accuracy is determined by calculating the log likelihood of each of the four options and selecting the highest-likelihood answer~\cite{eval-harness}.

We describe our language-specific prompt templates designed for our multilingual machine reading comprehension experiments. In order to maintain both a consistent structure across languages and the native labeling of input fields, we map the input fields (passage, question, and answer options) to their corresponding local terms. Table~\ref{tab:prompt_templates_belebele} provides a summary of the mappings for each target language along with the respective answer indicator. In our experiments, these templates are used to format the input as a single text string that is passed to the model.

\begingroup
  \microtypesetup{protrusion=false}
  \begin{CJK*}{UTF8}{gkai}
  \begin{table}[ht]
    \centering
    \footnotesize
    \begin{tabular}{lcccc}
      \toprule
      Language    & Passage & Question & Option Labels  & Answer Indicator  \\
      \midrule
      Chinese     & {\CJKfamily{gbsn} 文档}    & {\CJKfamily{gbsn} 问题}    & A, B, C, D     & {\CJKfamily{gbsn}答案}             \\
      German      & Dokument & Frage   & A, B, C, D     & Antwort          \\
      Korean      & 문서    & 질문    & 가), 나), 다), 라) & 정답             \\
      Indonesian  & Dokumen & Pertanyaan & A, B, C, D   & Jawaban          \\
      Swahili     & Hati    & Swali   & A, B, C, D     & Jibu             \\
      Yoruba      & Iwe     & Ibéèrè  & A, B, C, D     & Idahun           \\
      \bottomrule
    \end{tabular}
    \caption{Mapping of Input Fields and Answer Indicators for Each Language in the BELEBELE benchmark.}
    \label{tab:prompt_templates_belebele}
  \end{table}
  \end{CJK*}
\endgroup

\section{Analysis of Generation Prompt Strategy}
\label{appendix:add-zeroshot}

In our main performance evaluation on AlpacaEval (Table~\ref{tab:alpaca}), the models generated responses without including system prompts instructing them to answer in the respective target languages. According to the study by \citet{marchisio2024understanding}, adding instructions that direct the model to respond in the target language can alleviate language confusion. Therefore, we conducted additional experiments to assess the impact of such zero-shot prompts on the models' performance.

Specifically, we included an instruction in the system prompt, such as "Please answer in the same language as the input," translated into the target language. This was intended to encourage the model to generate responses in the appropriate language. We then performed AlpacaEval using these adjusted prompts, and the results are presented in Table~\ref{tab:zero-shot}.

As a result, we observed that even with the inclusion of zero-shot prompts instructing the models to respond in the target language, the overall trends remained similar to those without such prompts. While there were slight improvements in win rates for some models and languages, the performance gains were not substantial. This outcome highlights the limitations of zero-shot prompts in significantly enhancing the target language generation capabilities of the models.

These findings suggest that simply instructing models to answer in the same language as the input is insufficient for overcoming language generation challenges in multilingual contexts.

\begin{table*}[ht]
\centering
\resizebox{0.60\linewidth}{!}{
\begin{tabular}{lcccc}
\toprule
Model & Eval Language & Chinese & Korean & Swahili \\
\midrule
Llama-3-8B & Target & 86.1 & 76.5 & 70.8 \\
Llama-2-7B & Target & 64.7 & 54.5 & 64.1 \\
Llama-2-13B & Target & 62.0 & 53.4 & 74.4 \\
Mistral-7B-v0.1 & Target & 57.9 & 50.5 & 51.3 \\
\bottomrule
\end{tabular}
}
\caption{Win rates (\%) of \textbf{CLO over SFT} on AlpacaEval when models are prompted to answer in the same language as the input using zero-shot prompts.}
\label{tab:zero-shot}
\end{table*}

\section{Analysis of AlpacaEval Prompt Strategies}
\label{appendix:alpaca_prompt}
In our experiments using AlpacaEval (as shown in Table~\ref{tab:alpaca}), a comparative analysis between the language-specific prompts used for the target language and the AlpacaEval prompts, which show high correlation with human evaluators, is presented in Table~\ref{tab:alpaca_prompt_modified}.\footnote{Since the original AlpacaEval prompt was used for evaluating English performance, we focus solely on the evaluation of the target language using language-specific prompts.}

Apart from two cases in the generated outputs of the Llama-2-7B Chinese model, both prompt types resulted in identical performance, indicating that the language-specific prompts we used also demonstrate a high correlation with human evaluations. Additionally, a manual review revealed that discrepancies in rankings between the evaluations using language-specific prompts and the original AlpacaEval prompts were due to the presence of English entities mixed in the responses, which, when translated or rewritten in the target language, were deemed to be of higher quality. This finding aligns with our intentions, and thus, our main performance results are reported using the AlpacaEval evaluations with language-specific prompts.
\begin{table}[ht]
\centering
\resizebox{0.46\linewidth}{!}{
\begin{tabular}{|c|c|c|c|}
\hline
\textbf{Model}          & \textbf{Language} & \textbf{Language-specific} & \textbf{Original} \\ \hline
\multirow{3}{*}{Llama-2-7B}  
    & Chinese        & 61.1        & 61.2     \\ \cline{2-4} 
    & Korean         & 53.8        & 53.8     \\ \cline{2-4}
    & Swahili        & 65.0        & 65.0     \\ \hline
\multirow{3}{*}{Llama-2-13B} 
    & Chinese        & 65.2        & 65.2     \\ \cline{2-4} 
    & Korean         & 53.9        & 53.9     \\ \cline{2-4}
    & Swahili        & 70.9        & 70.9     \\ \hline
\multirow{3}{*}{Llama-3-8B}  
    & Chinese        & 83.0        & 83.0     \\ \cline{2-4}
    & Korean         & 77.8        & 77.8     \\ \cline{2-4}
    & Swahili        & 70.4        & 70.4     \\ \hline
\multirow{3}{*}{Mistral-7B-v0.1}  
    & Chinese        & 57.4        & 57.4     \\ \cline{2-4}
    & Korean         & 50.5        & 51.5     \\ \cline{2-4}
    & Swahili        & 51.3        & 51.3     \\ \hline
\end{tabular}
}
\caption{Comparison of performance between our language-specific prompts and the original prompts. The table shows the win rate (\%) compared to the models trained with SFT by language for each model based on the type of prompt used.}
\label{tab:alpaca_prompt_modified}
\end{table}

\newpage

\section{Analysis of MMMLU Performance in Target Languages}
\label{Appendix:MMMLU_analysis}

\begin{figure}[ht]
 \centering
\includegraphics[width=0.75\linewidth]{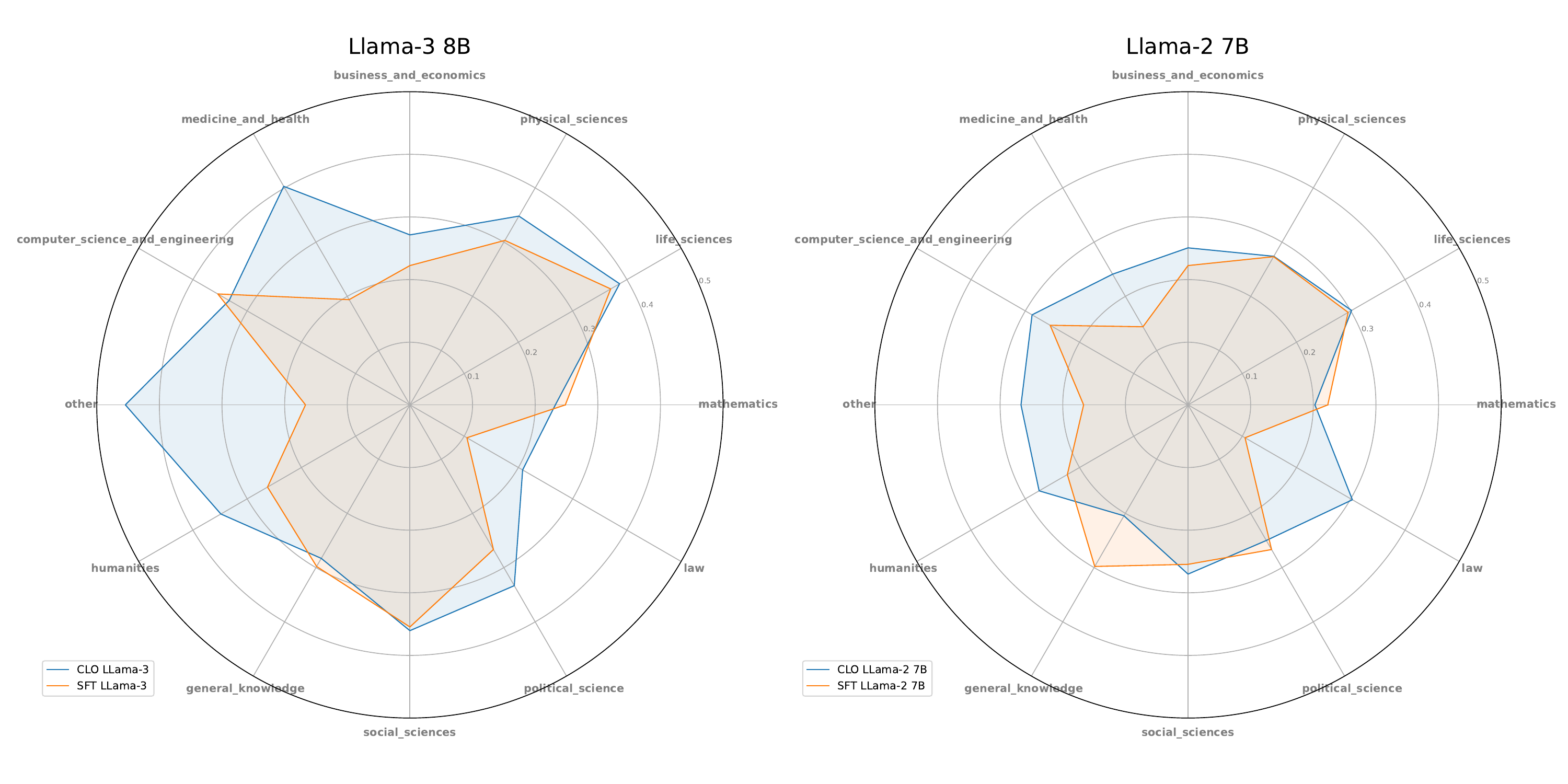}
\caption{Comparison of average MMMLU performance by category for CLO and SFT models of Llama-2 and Llama-3 in Chinese, Korean, and Swahili languages.}
 \label{fig:example}
\end{figure}

We present a comprehensive analysis of the MMMLU performance of our proposed CLO Llama-3 and Llama-2 7B model compared to the SFT Llama-3 and Llama-2 7B model across various subjects in Chinese, Korean, and Swahili, as illustrated in Figure~\ref{fig:example}. To conduct a more detailed analysis, we refined the existing 57 categories defined in MMMLU into 24 more specific categories.

\newpage

\section{Effect of Training Data Size on Llama-3}
\label{appendix:Llama3-data}
\begin{figure}[ht]
 \centering
\includegraphics[width=0.9\linewidth]
{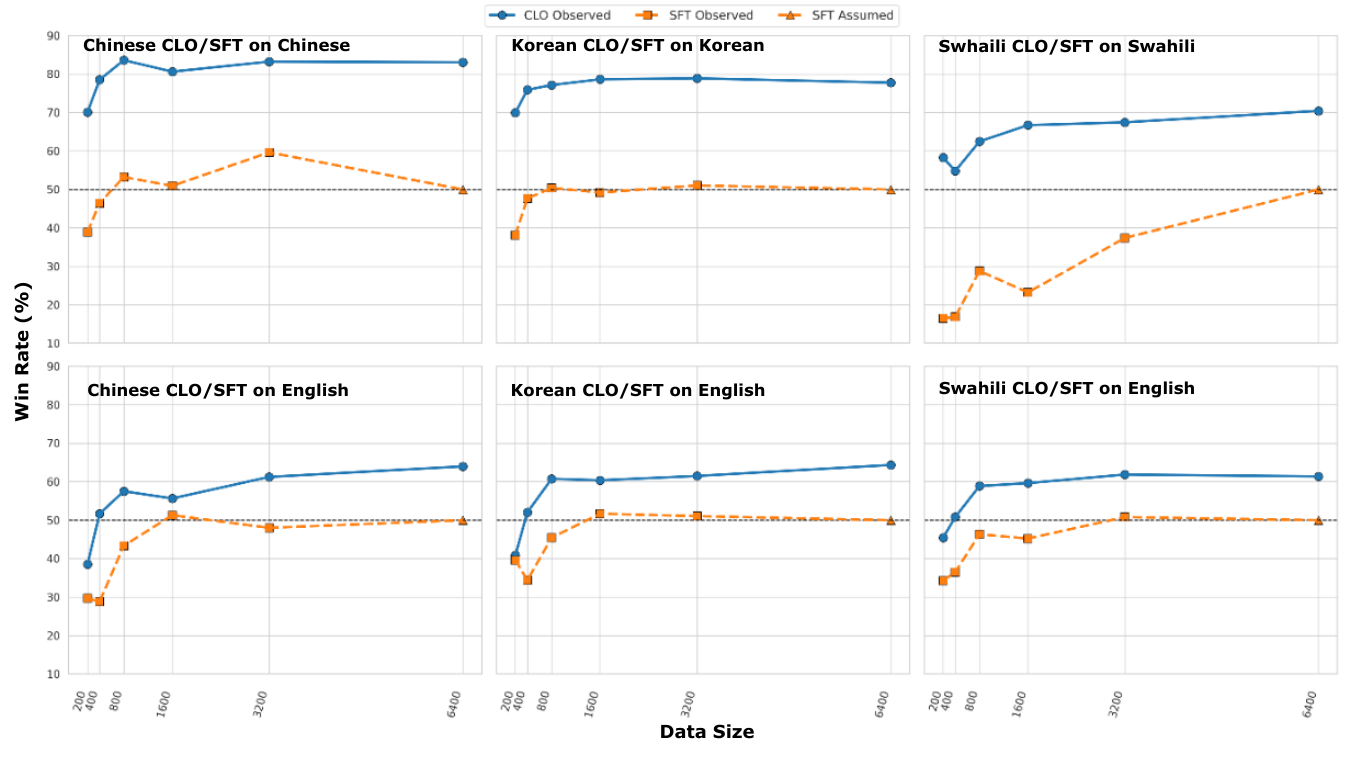}
\caption{Comparison of win rates between CLO and SFT Llama-3 models, trained with varying amounts of data, against a model fine-tuned using the SFT method with 6,400 pair examples on the AlpacaEval dataset. The "SFT Assumed" baseline is assigned a win rate of 50\% since it compares the same model and represents the ideal performance of an SFT model trained with fewer than 6,400 pairs.}
 \label{fig:dataperformance}
\end{figure}

\end{document}